\documentclass[letterpaper]{article} 
\usepackage{aaai2027}  
\nocopyright  
\usepackage[hyphens]{url}  
\usepackage{graphicx} 
\usepackage{natbib}  
\usepackage{caption} 
\usepackage{algorithm}
\usepackage{algorithmic}

\usepackage{booktabs}
\usepackage{tabularx}

\usepackage{amsmath, amssymb, amsfonts}
\usepackage{amsthm}
\usepackage{mathtools}
\usepackage{bm}
\usepackage{subcaption}

\usepackage{tikz}
\usetikzlibrary{arrows.meta, positioning, shapes.geometric, fit, backgrounds, calc}
\definecolor{dircol}{RGB}{30,80,200}    
\definecolor{spillcol}{RGB}{200,40,40}  
\definecolor{selcol}{RGB}{90,90,90}     
\definecolor{srcfill}{RGB}{225,235,252}  
\definecolor{tgtfill}{RGB}{237,237,237}  
\definecolor{trnfill}{RGB}{230,245,235}  

\newtheorem{assumption}{Assumption}
\newtheorem{definition}{Definition}
\newtheorem{theorem}{Theorem}

\newcommand{\indep}{\perp\!\!\!\perp}

\title{Transportable Causal Effect Estimation across Networks under Interference}
\author{
    Xiaojing Du\textsuperscript{\rm 1}, Jiuyong Li\textsuperscript{\rm 1},
    Lin Liu\textsuperscript{\rm 1}, Debo Cheng\textsuperscript{\rm 2},
    Jixue Liu\textsuperscript{\rm 1}, Thuc Duy Le\textsuperscript{\rm 1}
}
\affiliations{
    \textsuperscript{\rm 1}Adelaide University, Adelaide, Australia\\
    \textsuperscript{\rm 2}Hainan University, Haikou, China\\
    \texttt{xiaojing.du@adelaide.edu.au}; \texttt{jiuyong.li@adelaide.edu.au};
    \texttt{lin.liu@adelaide.edu.au}\\
    \texttt{jixue.liu@adelaide.edu.au}; \texttt{thuc.le@adelaide.edu.au};
    \texttt{chengd@hainanu.edu.cn}
}

\begin{document}

\maketitle

\begin{abstract}
Estimating causal effects under network interference typically assumes
that the network used for training and the network used for deployment
coincide. In practice, an intervention is run on one population while
the question of interest concerns a different population, and the two
generally differ in topology, node-covariate composition, and
spillover pathways. Transporting a causal effect across networks is
therefore a data-fusion problem that no existing algorithm solves. We employ a
selection diagram, extended to the network setting so that covariate
shift and structural network shift enter as separate selectors, and
derive from it a transport formula for the direct, spillover, and total
effects in the deployment population. Each formula makes explicit which
interventional mechanism is assumed invariant and which observational
distribution must be reweighted. We then turn the formulas into TranCE
(Transported Causal Effects), a doubly-robust algorithm combining an
interventional outcome model, a domain density-ratio correction, and
cross-fitted inference. Extensive experiments on two semi-synthetic
benchmarks derived from real-world social networks and on a fully real
weather-insurance field experiment, where the transported effects are
checked against held-out randomized estimates, confirm the
effectiveness of our approach. Our findings have the potential to improve
intervention strategies in networked systems, particularly in social
networks and public health.
\end{abstract}


\section{Introduction}
\label{sec:intro}

Estimating causal effects from networked data is central in social
science~\citep{forastiere2021identification}, public
health~\citep{hudgens2008toward}, online
platforms~\citep{eckles2017design}, and policy
evaluation~\citep{sobel2006randomized}. In these settings a unit's
outcome depends on its own treatment and covariates and, through
interference, on the treatments of its neighbours, so the same policy
can produce different effects when deployed on different networks.

That dependence is what makes an effect estimated on one network fail to
carry over to another. Most existing estimators are designed for a
same-network regime, where the graph used to identify, train, or
validate the estimator is also the graph on which the effect is
evaluated. Deployment routinely violates this. An intervention may be
feasible in one domain $\Pi$, but the policy question concerns a
different domain $\Pi^{*}$ where only observational data and the graph
structure are available. The two domains differ in node attributes,
degree distribution, and topology, so an estimator fit on $\Pi$ does not
automatically identify the causal effect in $\Pi^{*}$. A concrete case
makes the difficulty visible.

In Melbourne, the causal effect of a policy is estimated through a
randomized experiment in which participants are randomly assigned to
treatment and control groups, and their outcomes are subsequently
tracked and collected. When the same policy is implemented in Sydney,
however, the population has a different demographic distribution, and
its social network structure also differs. The government is interested
in estimating the policy's effect in Sydney without conducting another
randomized experiment.

This is a causal effect transportability problem on network data, and
no existing algorithm solves it. Single-network interference
methods~\citep{hudgens2008toward,aronow2017estimating,forastiere2021identification}
identify effects only on the graph the experiment was run on.
Networked causal
estimators~\citep{ma2021causal,jiang2022estimating,du2025telling,wu2025causal}
assume the deployment graph equals the estimation graph, so a fitted
model carries no validity condition elsewhere. Treatment-effect
transfer and trial
generalization~\citep{kunzel2018transfer,dahabreh2020extending,bica2022transfer}
move effects between populations but treat units as independent, which
fails exactly when a unit responds to its neighbours' treatments.
Graph-domain generalization~\citep{wu2022handling} trains
representations to be stable across graphs, a predictive criterion, and
carrying that machinery over to effect
estimation~\citep{sui2024invariant} delivers invariance of the
representation rather than an identified causal effect in the new
domain. What is missing is an algorithm that takes an experiment on one
network and an observed second network and returns the causal effects
on the second, and we present an algorithm to fill in this gap.

In this paper, we present TranCE (Transported Causal Effects), an
algorithm for transportable causal effect estimation on network data.
The two discrepancies in the example above are modelled separately:
covariate shift changes the distribution of node attributes, while
structural network shift changes graph-derived quantities such as
neighbourhood composition and structural position. We encode both in a
selection diagram extended to the network setting, derive from it
transport formulas for the direct, spillover, and total effects in the
deployment domain, and estimate the formulas with a doubly robust
combination of an interventional outcome model and a domain
density-ratio correction, with cross-fitted inference.

Our contributions are threefold.
\begin{itemize}
\item We establish the graphical conditions under which causal effects
transport across networks under interference, employing a selection
diagram extended to the network setting so that covariate shift and
structural network shift enter as separate selectors.

\item We derive the transport formulas for the direct, spillover, and
total effects in $\Pi^{*}$, which state exactly which interventional
information from $\Pi$ is reused and which observational information
from $\Pi^{*}$ is required for adjustment.

\item We develop TranCE (Transported Causal Effects), a doubly robust
algorithm that implements those formulas with an interventional outcome
model, a domain density-ratio correction, and known neighbour treatment
propensities, and we evaluate it on real cross-network benchmarks and a
fully real field experiment.
\end{itemize}

\section{Related Work}
\label{sec:related}

This section places our setting against four lines of work, none of
which identifies a causal effect in $\Pi^{*}$ from interventions in
$\Pi$ and observations in $\Pi^{*}$.

\paragraph{Causal inference under network interference.}
Classical work formalizes how one unit's treatment affects another's
outcome and develops estimands under partial interference and exposure
mappings~\citep{hudgens2008toward,sobel2006randomized,aronow2017estimating},
with identification results for observational network
studies~\citep{forastiere2021identification} and designs for network
experiments~\citep{eckles2017design}. These are same-network results:
the graph that defines a unit's neighbours is also the one on which the
estimand is evaluated.

\paragraph{Graph-based causal estimation on networks.}
Graph-based estimators use representation learning to model network
structure, latent confounding, and interference, spanning neural
treatment-effect models~\citep{shi2019adapting}, networked and hypergraph
interference~\citep{ma2021causal,jiang2022estimating,ma2022learning},
generalization analysis~\citep{cai2023generalization}, targeted
learning~\citep{chen2024doubly}, and graph transformers for unknown
interference~\citep{wu2025causal}. They improve within-network
estimation without giving graphical conditions under which an effect
learned on one graph is identified on a different graph.

\paragraph{Transportability, treatment-effect transfer, and graph-domain generalization.}
Causal transportability studies when causal knowledge transfers across
environments using selection diagrams and
do-calculus~\citep{pearl2011transportability,bareinboim2016causal}, and
trial generalization combines experimental information from one domain
with observational covariates from
another~\citep{kunzel2018transfer,dahabreh2020extending,bica2022transfer,wei2023transfer,sun2023treatment}.
These are directly relevant to our two-domain setting but are tabular or
population-level, without interference. A separate line studies
graph-domain generalization and invariant graph
learning~\citep{wu2022handling,chen2022learning,li2022ood,wu2024graph},
within which \citet{sui2024invariant} extend invariant learning to
causal effect estimation on graphs, giving representation-level
invariance rather than graphical identification in $\Pi^{*}$. Closest to
us, \citet{hoshino2025transfer} transfers policy effects under
interference when the deployment network is unobserved, which permits
only partial identification and precludes adjustment using the realized
topology. In our setting the observed graph of $\Pi^{*}$ is part of the
identifying functional, since it determines the structural features,
neighbour summaries, and neighbour treatment levels in $\Pi^{*}$.

\section{Transportability Framework}
\label{sec:theory}

This section establishes the formal framework for transporting causal
effects across networks. We introduce notation and define three
causal estimands that capture distinct sources of treatment variation
under network interference.
We then introduce network selection diagrams as the representational
language for encoding distributional differences between domains,
state and prove the transportability theorem, and finally derive a
doubly-robust estimator for the transported functional.

\subsection{Setup and Notation}
\label{sec:notation}

Consider two domains $\Pi$ and $\Pi^{*}$ with distinct network
structures $\mathcal{G} = (\mathcal{V}, \mathcal{E})$ and
$\mathcal{G}^{*} = (\mathcal{V}^{*}, \mathcal{E}^{*})$. Interventional
data are available in $\Pi$, while only passive observations are
available in $\Pi^{*}$. Each node $i$ carries a binary treatment
$T_i \in \{0, 1\}$, pre-treatment covariates
$\bm{Z}_i \in \mathbb{R}^d$, and an outcome $Y_i \in \mathbb{R}$.
We write $\mathcal{N}(i)$ for the neighbourhood of $i$. The
interventional outcome of node $i$ depends on its own treatment and
its neighbours' treatments:
$P\big(y_i \mid \mathrm{do}(t_i, \bm{t}_{\mathcal{N}(i)})\big)$. Here
$\mathrm{do}(\cdot)$ denotes an intervention that sets a variable by
external assignment rather than by observation, so this expression is
the outcome distribution when node $i$ and its neighbours are assigned
the stated treatments.

The \emph{structural positional features}
$\bm{F}_i^{\mathrm{str}}$ summarise the network around node $i$ and are
a function of the topology alone, in our implementation the degree, the
mean-normalised degree, and the log degree. The \emph{neighbour
covariate aggregate} $\bar{\bm{Z}}_i = \frac{1}{|\mathcal{N}(i)|}
\sum_{j \in \mathcal{N}(i)} \bm{Z}_j$ summarises the demographics of
those neighbours, as $\bm{Z}_i$ does for the node itself. The three are
pre-treatment and together form the context
$\bm{w}_i=(\bm{Z}_i,\bm{F}_i^{\mathrm{str}},\bar{\bm{Z}}_i)$.

The third quantity is the \emph{neighbour treatment} $E_i$, which takes
values in a finite set $\mathcal B$ and records how much treatment
reaches node $i$ through its neighbours. It is the second treatment
variable of the model: node $i$ responds to its own treatment along
$T_i \to Y_i$ and to that of its neighbours along $E_i \to Y_i$, and the
estimands of Section~\ref{sec:effects} intervene on both.

The neighbour treatment is derived from the realized treatments of the
neighbours together with the degree: writing
$\bar{T}_{\mathcal{N}(i)} = \frac{1}{|\mathcal{N}(i)|}
\sum_{j \in \mathcal{N}(i)} T_j$ for the fraction of treated
neighbours, $E_i = b(\bar{T}_{\mathcal{N}(i)}, |\mathcal N(i)|)$ sends
the fraction and the degree to one of finitely many levels, and our
implementation cuts $\bar{T}_{\mathcal{N}(i)}$ at its terciles into
low, medium, and high.

Identification rests on $E_i$ being a sufficient summary of the
neighbours' treatments, the stratified-interference condition standard
in that
literature~\citep{hudgens2008toward,aronow2017estimating,forastiere2021identification}:
conditional on $T_i$, $E_i$, and $\bm{w}_i$, the outcome $Y_i$ does
not depend on which particular neighbours are treated, so the whole
neighbourhood treatment vector can be replaced by the single level
$E_i$.

We also require an overlap condition, which secures both the
identification of Section~\ref{sec:transport} and the estimator of
Section~\ref{sec:dr}.
\begin{assumption}[Overlap]
\label{ass:overlap}
There exists $\delta\in(0,1)$ such that, for $P^{*}$-almost every
context $\bm{w}=(\bm{Z}_i,\bm{F}_i^{\mathrm{str}},\bar{\bm{Z}}_i)$,
a unit with context $\bm{w}$ arises in the interventional domain
$\Pi$ with probability at least $\delta$, and within $\Pi$ each
own-treatment level and neighbour treatment level of interest occurs with
probability at least $\delta$ given $\bm{w}$.
\end{assumption}
Both parts are needed: a context that occurs in $\Pi^{*}$ but never in
$\Pi$ reports no experimental outcome, and a pair $(t,e)$ never
realized in $\Pi$ leaves $\mu$ unobserved there, so the contrast that
defines the effect cannot be formed.

\subsection{Effect Definitions and Problem Formulation}
\label{sec:effects}

Write $t,t'$ for the baseline and intervened levels of a node's own
treatment and $e,e'$ for two neighbour treatment levels.

\begin{definition}[Direct Effect]
$\tau_i^{\mathrm{dir}} = \mathbb{E}[Y_i \mid \mathrm{do}(t', e)]
 - \mathbb{E}[Y_i \mid \mathrm{do}(t, e)]$, the change in node $i$'s
outcome from intervening on its own treatment while its neighbours stay
at $e$.
\end{definition}

\begin{definition}[Spillover Effect]
$\tau_i^{\mathrm{spill}} = \mathbb{E}[Y_i \mid \mathrm{do}(t, e')]
 - \mathbb{E}[Y_i \mid \mathrm{do}(t, e)]$, the change from
intervening on the neighbour treatment while node $i$'s own treatment
stays at $t$.
\end{definition}

\begin{definition}[Total Effect]
$\tau_i^{\mathrm{total}} = \mathbb{E}[Y_i \mid \mathrm{do}(t', e')]
 - \mathbb{E}[Y_i \mid \mathrm{do}(t, e)]$, the joint change from
intervening on both.
\end{definition}

\paragraph{Problem formulation.}
Domain $\Pi$ has network $\mathcal{G}$ and demographic distribution
$P$, an experiment is conducted in it, and
$\tau^{\mathrm{dir}}, \tau^{\mathrm{spill}}, \tau^{\mathrm{total}}$
are obtained from that experiment. Domain $\Pi^{*}$ has network
$\mathcal{G}^{*}$ and demographic distribution $P^{*}$, and only
observational data are available in it. We apply the same treatment to
$\Pi^{*}$ and estimate $\mathbb{E}^{*}[\tau_i^{\mathrm{dir}}]$,
$\mathbb{E}^{*}[\tau_i^{\mathrm{spill}}]$, and
$\mathbb{E}^{*}[\tau_i^{\mathrm{total}}]$ over $i \in \mathcal{V}^{*}$
without conducting an experiment there.

Two obstacles make this nontrivial. First, $\mathcal{G}^{*}$ differs
from $\mathcal{G}$ in topology, connectivity patterns, and degree
distribution, which directly affects interference pathways. Second, the
distribution of covariates in $\Pi^{*}$ differs from that in $\Pi$,
altering the causal effects of treatment on the outcome.

\subsection{Network Selection Diagrams}
\label{sec:diagrams}

\begin{figure}[t]
  \centering
  \resizebox{0.92\columnwidth}{!}{%
  \begin{tikzpicture}[>={Latex[length=1.9mm]}, font=\small, yscale=0.80,
    v/.style={circle,draw,minimum size=8.5mm,inner sep=0pt},
    s/.style={rectangle,draw=selcol,thick,fill=selcol,minimum size=5.5mm,inner sep=1pt,font=\small\bfseries,text=white},
    e/.style={->,draw,gray!85},
    sel/.style={->,draw=selcol,thick,dashed}]
    \node[s] (SG) at (-3.7,0.2)  {$S_G$};
    \node[v] (F)  at (-2.0,0.4)  {$\bm F_i^{\mathrm{str}}$};
    \node[v] (Zb) at (-0.3,2.0)  {$\bar{\bm Z}_i$};
    \node[v] (E)  at (-1.2,-2.05) {$E_i$};
    \node[v] (Y)  at (1.15,0.3)  {$Y_i$};
    \node[v] (T)  at (2.9,1.15)  {$T_i$};
    \node[v] (Z)  at (3.1,-1.55) {$\bm Z_i$};
    \node[s] (SZ) at (4.6,1.15)  {$S_Z$};
    \draw[sel] (SG) -- (F);
    \draw[sel] (SG) to[bend left=16] (Zb);
    \draw[sel] (SG) to[bend right=14] (E);
    \draw[sel] (SZ) -- (Z);
    \draw[sel] (SZ) -- (T);
    \draw[e] (F) -- (E);
    \draw[e] (F) to[bend right=3] (Y);  \draw[e] (F) to[bend left=8] (T);
    \draw[e] (Zb) -- (E);  \draw[e] (Zb) to[bend left=14] (Y);  \draw[e] (Zb) to[bend left=6] (T);
    \draw[e] (Z) -- (Y);    \draw[e] (Z) -- (T);
    \draw[e] (Z) to[bend right=6] (E);
    \draw[->,dircol,very thick] (T) -- (Y);
    \draw[->,spillcol,very thick] (E) to[bend right=18] (Y);
  \end{tikzpicture}}
  \caption{\textbf{Network selection diagram $\mathcal D$.} The blue
  arrow is the direct pathway $T_i\to Y_i$ from the node's own treatment
  and the red arrow is the spillover pathway $E_i\to Y_i$ from the
  treatment of its neighbours. The grey squares are the selection
  variables, covariate shift $S_Z$ and structural shift $S_G$, and each
  dashed arrow points to a variable whose generating mechanism changes
  across $\Pi$ and $\Pi^{*}$.}
  \label{fig:selection-diagram}
\end{figure}
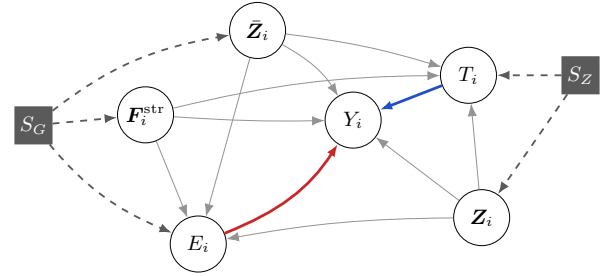

Transporting an effect requires knowing which parts of the data
generating process the two domains share, and a selection diagram is
the device that records exactly that. We first recall the original
definition and then extend it to the network setting.

\begin{definition}[Selection Diagram; \citealp{pearl2011transportability}]
\label{def:pb-diagram}
Let $\mathcal{D}_0$ be the causal diagram shared by two domains. A
selection diagram is $\mathcal{D}_0$ augmented with a set $S$ of
selection variables, drawn as square nodes, where $S_V \to V$ is present
whenever the mechanism that generates $V$ may differ between the two
domains. A variable with no incoming selection arrow has the same
mechanism in both.
\end{definition}

Our setting adds structure to this: the two domains differ through two
distinct channels, who the units are and how they are connected, and
these two channels act on disjoint sets of variables.

\begin{definition}[Network Selection Diagram]
\label{def:diagram}
A network selection diagram $\mathcal{D}$ is a selection diagram in the
sense of Definition~\ref{def:pb-diagram} over the node variables
$\{\bm{Z}_i, T_i, \bm{F}_i^{\mathrm{str}},
\bar{\bm{Z}}_i, E_i, Y_i\}_{i \in \mathcal{V}}$ with exactly two
selection variables. The selector $S_Z \to (\bm{Z}_i, T_i)$ encodes
covariate shift, meaning $P(\bm{Z}) \neq P^{*}(\bm{Z})$ together
with the change of assignment mechanism from a randomized $\Pi$ to an
observational $\Pi^{*}$. The selector
$S_G \to (\bm{F}_i^{\mathrm{str}}, \bar{\bm{Z}}_i, E_i)$
encodes network structural shift, meaning
$\mathcal{G} \neq \mathcal{G}^{*}$, reflecting that a change in graph
topology simultaneously alters node $i$'s structural features, its
neighbour covariate aggregate, and its neighbour treatment level.
\end{definition}

On notation, we write $P^{*}(\cdot) = P(\cdot \mid s^{*})$ for the
distribution in $\Pi^{*}$, where $s^{*}$ is the configuration of the
selection variables.

\subsection{Transportability of Network Causal Effects}
\label{sec:transport}

Figure~\ref{fig:selection-diagram} shows the network selection diagram
used in this paper. Our notion of transportability builds on the
transportability selection diagram of \citet{pearl2011transportability}
and applies it to the network-interference setting.

\begin{definition}[Transportability under Network Interference]
\label{def:transport}
Given $\Pi$, $\Pi^{*}$, and selection diagram $\mathcal{D}$, a causal
effect
$\tau^{R}$ with $R \in \{\mathrm{dir}, \mathrm{spill}, \mathrm{total}\}$
is transportable from $\Pi$ to $\Pi^{*}$ if $\mathbb{E}^{*}[\tau^{R}_i]$ can
be expressed as a functional of (i) $S$-free interventional quantities
estimable from $\Pi$, and (ii) do-free observational quantities
estimable from $\Pi^{*}$, in which $S$ appears only as a conditioning
variable.
\end{definition}

Define the \emph{interventional mechanism function}
\begin{equation*}
\begin{aligned}
\mu(t, e, \bm{z}, \bm{f}, \bar{\bm{z}}) := \mathbb{E}\big[Y_i \bigm|
  &\ \mathrm{do}(t, e),\ \bm{Z}_i = \bm{z}, \\
  &\ \bm{F}_i^{\mathrm{str}} = \bm{f},\
    \bar{\bm{Z}}_i = \bar{\bm{z}}\big],
\end{aligned}
\end{equation*}
identifiable from $\Pi$'s interventional data together with its network
and demographic measurements.

\begin{theorem}[Transportability of Network Causal Effects]
\label{thm:transport}
Let $\mathcal{D}$ be the network selection diagram for $\Pi$ and
$\Pi^{*}$ as in Figure~\ref{fig:selection-diagram}, and suppose the stratified-interference condition of
Section~\ref{sec:notation} and the overlap condition of
Assumption~\ref{ass:overlap} hold. The three
effects $\tau^{\mathrm{dir}}$, $\tau^{\mathrm{spill}}$,
$\tau^{\mathrm{total}}$ are transportable from $\Pi$ to $\Pi^{*}$ if
$\big( Y_i \indep \{S_Z, S_G\} \,\big|\,
\bm{Z}_i, \bm{F}_i^{\mathrm{str}}, \bar{\bm{Z}}_i
\big)_{\mathcal{D}_{\overline{T,E}}}$.
Here $\mathcal{D}_{\overline{T,E}}$ is the diagram $\mathcal{D}$ with
the incoming edges to $T_i$ and $E_i$ removed. The transport formulas
are
\begin{align}
\mathbb{E}^{*}[\tau_i^{\mathrm{dir}}]
&= \iiint \big[\, \mu(t', e, \bm{z}, \bm{f}, \bar{\bm{z}}) \notag \\
&\qquad - \mu(t, e, \bm{z}, \bm{f}, \bar{\bm{z}}) \,\big]\,
   dP^{*}(\bm{z}, \bm{f}, \bar{\bm{z}}),
   \label{eq:transport-dir} \\
\mathbb{E}^{*}[\tau_i^{\mathrm{spill}}]
&= \iiint \big[\, \mu(t, e', \bm{z}, \bm{f}, \bar{\bm{z}}) \notag \\
&\qquad - \mu(t, e, \bm{z}, \bm{f}, \bar{\bm{z}}) \,\big]\,
   dP^{*}(\bm{z}, \bm{f}, \bar{\bm{z}}),
   \label{eq:transport-spill} \\
\mathbb{E}^{*}[\tau_i^{\mathrm{total}}]
&= \iiint \big[\, \mu(t', e', \bm{z}, \bm{f}, \bar{\bm{z}}) \notag \\
&\qquad - \mu(t, e, \bm{z}, \bm{f}, \bar{\bm{z}}) \,\big]\,
   dP^{*}(\bm{z}, \bm{f}, \bar{\bm{z}}),
   \label{eq:transport-total}
\end{align}
\end{theorem}

The proof is given in full in Appendix~D.

\subsection{A Doubly-Robust Transport Estimator}
\label{sec:dr}

Theorem~\ref{thm:transport} identifies the target as
$\mathbb{E}^{*}[Y_i\mid \mathrm{do}(t,e)] = \int \mu(t,e,\bm{w})\,
dP^{*}(\bm{w})$. There are two classical routes to such a mean: fit
an outcome model and substitute it, unbiased when the model is correct,
or reweight the observed outcomes by the inverse probability of the
treatment received, unbiased when those probabilities are correct.
Substitution alone is fragile, since its first-order bias is the
average outcome-model error $\int (\hat\mu-\mu)\, dP^{*}$ and nothing
cancels it. We therefore build a one-step doubly-robust estimator that
combines the two routes, is consistent whenever \emph{either} is
correctly specified, and admits a $\sqrt{n^{*}}$ central limit theorem
under the product-rate condition of Theorem~\ref{thm:inference}.

\paragraph{What each domain supplies.}
The interventional domain $\Pi$ supplies $n$ nodes with covariates, own
treatment, neighbour treatment level, and outcome,
$\{(\bm w_i, T_i, E_i, Y_i)\}_{i=1}^{n}$, and its assignment is
randomized by design. The target $\Pi^{*}$ supplies $n^{*}$ nodes with
covariates and graph structure, $\{\bm w_j\}_{j=1}^{n^{*}}$. Write
$\pi_{S}(\bm{w})$ for the probability that a node with context
$\bm{w}$ comes from $\Pi$ rather than from $\Pi^{*}$, estimated from
the covariates of the two samples and their domain labels. Write $A_i := (T_i, E_i)$ for the
treatment pair of a source unit and
$\pi_{A}(t,e\mid \bm{w}) := \Pr(T = t, E = e \mid \bm{w})$,
within $\Pi$, for the joint treatment propensity.

\begin{theorem}[Doubly-Robust Transport]
\label{thm:dr}
Under Theorem~\ref{thm:transport} and Assumption~\ref{ass:overlap},
estimate the transported spillover effect of
Eq.~\eqref{eq:transport-spill} by
\begin{equation}
\label{eq:dr-est}
\begin{aligned}
\widehat{\mathbb E^{*}}[\tau^{\mathrm{spill}}]
  =\; &\tfrac{1}{n^{*}}\!\sum_{j \in \mathcal V^{*}}\!
        \big[\hat\mu(t,e', \bm{w}_j) - \hat\mu(t,e, \bm{w}_j)\big] \\
   +\;&\tfrac{1}{n^{*}}\!\!\sum_{\substack{i \in \mathcal V\\ A_i = (t,e')}}\!\!
        \tfrac{1-\hat\pi_{S}(\bm{w}_i)}
              {\hat\pi_{S}(\bm{w}_i)\,\hat\pi_{A}(t,e'\mid \bm{w}_i)}\,
        \big[Y_i - \hat\mu(t,e',\bm{w}_i)\big] \\
   -\;&\tfrac{1}{n^{*}}\!\!\sum_{\substack{i \in \mathcal V\\ A_i = (t,e)}}\!\!
        \tfrac{1-\hat\pi_{S}(\bm{w}_i)}
              {\hat\pi_{S}(\bm{w}_i)\,\hat\pi_{A}(t,e\mid \bm{w}_i)}\,
        \big[Y_i - \hat\mu(t,e,\bm{w}_i)\big],
\end{aligned}
\end{equation}
The estimator is consistent for the
transported effect of Theorem~\ref{thm:transport} if \emph{either}
$\hat\mu \to \mu$ \emph{or}
$(\hat\pi_{S}, \hat\pi_{A}) \to (\pi_{S}, \pi_{A})$ in $L^{2}$.
\end{theorem}

All three sums are divided by $n^{*}$ because all three estimate an
average over the target population. The first runs over the target
nodes, so it is the empirical version of the transport integral
$\iiint[\,\cdot\,]\,dP^{*}$ of Eq.~\eqref{eq:transport-spill}. The
other two run over the source nodes, where the odds factor
$(1-\hat\pi_S)/\hat\pi_S$ re-weights each residual so that the source
sample stands in for the target one. The direct and total estimators are identical
with the pair $\{(t,e'),(t,e)\}$ replaced by $\{(t',e),(t,e)\}$ and by
$\{(t',e'),(t,e)\}$ respectively.

\begin{theorem}[Asymptotic Normality and Valid Intervals]
\label{thm:inference}
Assume the conditions of Theorem~\ref{thm:dr}, and in addition that the
nuisances are cross-fitted, that the product of their $L^{2}$ errors is
$o_{p}(n^{*-1/2})$, and that sampling is in a limited-dependence regime.
Then
$\sqrt{n^{*}}\big(\widehat{\mathbb E^{*}}[\tau^{R}] -
\mathbb E^{*}[\tau^{R}]\big)$ is asymptotically normal with a variance
consistently estimated by $\widehat V^{*}$ of
Section~\ref{sec:crossfit}, so the Wald interval
$\widehat{\mathbb E^{*}}[\tau^{R}]\pm 1.96\sqrt{\widehat V^{*}}$ is
asymptotically valid. The estimator is also semiparametrically
efficient, attaining the variance bound of the influence function in
Appendix~D.
\end{theorem}

We defer both proofs to Appendix~D.

\section{Method}
\label{sec:method}

This section turns the transport formula of
Section~\ref{sec:theory} into a concrete estimator.
Theorems~\ref{thm:transport} and~\ref{thm:dr} reduce it to fitting
three nuisance functions, each on the data that identifies it. We
combine these components so the estimator stays consistent when either
the outcome model or the reweighting is correct.

The three nuisances differ in which domain they are read from.
i) The interventional outcome regression $\mu(t,e,\bm w)$ is fitted
on $\Pi$ alone, since it needs treatments and outcomes and only $\Pi$
has them. ii) The selection propensity
$\pi_S(\bm w)=\Pr(S=\Pi\mid\bm w)$, equivalently the domain
density ratio $r(\bm w)=(1-\pi_S(\bm w))/\pi_S(\bm w)$, is
the one quantity fitted on both domains, using their covariates and
domain labels and neither domain's treatments. iii) The joint treatment
propensity $\pi_A(t,e\mid\bm w)$ is known by design in $\Pi$. We
describe each below, then
substitute the estimates into Eq.~\eqref{eq:dr-est} to form the
doubly-robust transport estimator and give its cross-fitted inference.

\subsection{Interventional Outcome Regression $\hat\mu$}
\label{sec:mu-net}

The outcome regression is the substitution branch of
Eq.~\eqref{eq:dr-est}. Because $\Pi$ is randomized, its observational
regression of $Y$ on $(T,E,\bm w)$ already equals the interventional
mechanism $\mu(t,e,\bm w)$, so fitting the loss below on $\Pi$
estimates the target functional directly. The model predicts the
treatment-pair contrasts at every target node, with
$\hat\mu_\theta(t,e,\bm w_i)
= (1-t)\,h^{0}_{\mathrm{dir}}(\bm z_i)
+ t\,h^{1}_{\mathrm{dir}}(\bm z_i)
+ f_{\mathrm{spill}}\!\big(e,
\mathrm{GCN}_\theta(\bm Z, \mathcal G)_i\big)$,
where $h^{0}_{\mathrm{dir}}$ and $h^{1}_{\mathrm{dir}}$ are feed-forward
potential-outcome heads over the covariates, and $f_{\mathrm{spill}}$
reads the neighbour treatment level, encoded as an indicator over
$\mathcal B$, together with the node's graph convolutional network (GCN)
embedding. That embedding summarises the structural context
$(\bm F^{\mathrm{str}},\bar{\bm Z})$, so the adjustment set
enters $\hat\mu$ through message passing, while the selection propensity
conditions on the raw $\bm w$, keeping the weighting branch exact.
One parameter set $\theta$ is shared, message-passing on $\mathcal G$ and
on $\mathcal G^{*}$ to form the target embeddings the substitution term
reads. The regression loss is
\begin{equation}
\label{eq:loss-mu}
\mathcal L_\mu(\theta)
= \tfrac{1}{n}\!\sum_{i:S_i=\Pi}\!
   \big(Y_i - \hat\mu_\theta(T_i,E_i,\bm w_i)\big)^{2}.
\end{equation}

\subsection{Selection Propensity $\hat\pi_S$ and Density Ratio $\hat r$}
\label{sec:rnet}

The density ratio $\hat r$ is the reweighting branch of
Eq.~\eqref{eq:dr-est}: it carries the source residuals onto the target
covariate law, correcting what the fitted outcome model misses under
covariate and structural shift. We obtain it from the selection propensity $\pi_S$ of
Section~\ref{sec:dr}, implemented as a logistic regression with both
samples entering at their natural sizes $n$ and $n^{*}$, which is what
makes the $1/n^{*}$ normalisation of Eq.~\eqref{eq:dr-est} the correct
one, as Appendix~C shows. Because $\bm w$ is fully observed in both domains, no
missing-data correction is needed. The density ratio is
$\hat r(\bm w) = (1-\hat\pi_S(\bm w))/\hat\pi_S(\bm w)$,
clipped to a bounded range, with the
numeric range in Appendix~C. Clipping stabilizes the finite-sample weights without guaranteeing
population overlap, so we diagnose overlap empirically and read heavily
clipped estimates as regularized approximations. Logistic regression is
a deliberate low-variance choice here, since $\hat r$ enters the
correction multiplicatively and its variance inflates the estimator's
own.

\subsection{Treatment Propensity $\hat\pi_A$}
\label{sec:anet}

The joint treatment propensity prices how likely a source unit was to
occupy a contrasted treatment pair, turning its reweighted residual
into an unbiased within-pair correction in Eq.~\eqref{eq:dr-est}. In all our
settings $\Pi$ is a randomized experiment, so $\pi_A$ is known by design
and factorizes as
$\pi_A(t,e\mid \bm w) = \pi_T(t)\,\pi_E(e\mid t, \bm w)$, with
$\pi_T$ the design assignment probability and $\pi_E$ induced by the
neighbours' independent random assignment, available in closed form as a
binomial in the degree.

\subsection{Stabilisation, Cross-Fitting, and Inference}
\label{sec:crossfit}

\paragraph{Stabilising the correction.}
The correction is a sum of inverse-propensity-weighted residuals,
unbiased but noisy, so we stabilise it in two standard ways. Within each pair we
self-normalise the weights, the H\'ajek ratio
form~\citep{hajek1971comment}, which divides the weighted residual sum
by the weight sum rather than by $n^{*}$. We then apply a positive-part
shrinkage that keeps the full adjustment when the correction is
supported by its own standard error and falls back to the substitution
when it is indistinguishable from zero. Both are asymptotically
equivalent to the raw one-step estimator of Theorem~\ref{thm:dr}.

\paragraph{Cross-fitting and inference.}
To obtain valid $\sqrt{n^{*}}$ inference with neural nuisance estimators,
we use $K$-fold cross-fitting~\citep{chernozhukov2018double}. Each
domain's index set is split into $K$ folds, the nuisances
$(\hat\mu^{(-\ell)}, \hat r^{(-\ell)}, \hat\pi_A^{(-\ell)})$ are
trained with fold $\ell$ of both domains held out, and the DR term is
scored on the held-out fold, which keeps any observation out of the fit
that scores it and renders the nuisance remainder second order once the
product of the nuisance errors is $o_p(n^{*-1/2})$, the condition of
Theorem~\ref{thm:inference}. The fold-averaged
variance
$\widehat V^{*}=n^{*-2}\big[\sum_{j\in\mathcal V^{*}}(\hat g_j-\bar g)^2
+\sum_{i\in\mathcal V}\widehat{\mathrm{corr}}_i^{2}\big]$
supplies the Wald interval
$\widehat{\mathbb E^{*}}[\tau^{R}] \pm 1.96\sqrt{\widehat V^{*}}$
\citep{vandervaart1998asymptotic}, valid under
Theorem~\ref{thm:inference}.

\section{Experiments}

\label{sec:exp}

We evaluate TranCE on two semi-synthetic cross-network benchmarks and a
real field experiment. On the semi-synthetic benchmarks, we compare
transported direct, spillover, and total effects with network-effect and
domain-transfer baselines over rotating source-target pairs. We then
conduct component ablations, followed by stress tests for
outcome-model misspecification, structural discrepancy, and hidden
confounding. Finally, we perform leave-region-out validation on the
field experiment.

\subsection{Datasets}

We adopt two real-world social-network families, where a node is a user
and an edge is a friendship: Twitch-Explicit
\citep{rozemberczki2021multi} and Facebook-100
\citep{traud2012social}. For both
families the raw node features are mapped to 50 dimensions, and the
treatment and neighbour-mediated outcome are semi-synthetic with the
generative details in Appendix~C. Our third
dataset is the Cai insurance experiment~\citep{cai2015social}, a fully
real weather-insurance field experiment in rural China with a household
friendship network, a randomized information-session treatment, and
observed insurance take-up across three separated regions. The source
code, the data-preparation scripts, and the instructions needed to
reproduce every number below are included in the code and data
supplement accompanying this submission.

\begin{table*}[!t]
\centering

\small
\setlength{\tabcolsep}{4pt}
\begin{tabular}{lcccccc}
\toprule
 & \multicolumn{3}{c}{Twitch} & \multicolumn{3}{c}{Facebook-100} \\
\cmidrule(lr){2-4}\cmidrule(lr){5-7}
Method & dir & spill & total & dir & spill & total \\
\midrule
TARNet & $0.084_{\pm 0.075}$ & $1.290_{\pm 0.018}$ & $1.281_{\pm 0.114}$ & $0.080_{\pm 0.064}$ & $1.411_{\pm 0.001}$ & $1.437_{\pm 0.099}$ \\
IPW & $0.523_{\pm 0.050}$ & $1.202_{\pm 0.050}$ & $1.726_{\pm 0.091}$ & $0.512_{\pm 0.138}$ & $1.317_{\pm 0.114}$ & $1.837_{\pm 0.246}$ \\
IGL & $0.094_{\pm 0.097}$ & $0.633_{\pm 0.248}$ & $0.713_{\pm 0.321}$ & $0.199_{\pm 0.162}$ & $1.009_{\pm 0.179}$ & $1.208_{\pm 0.335}$ \\
Hoshino & $0.104_{\pm 0.029}$ & $0.780_{\pm 0.139}$ & $0.830_{\pm 0.142}$ & $0.114_{\pm 0.028}$ & $1.018_{\pm 0.030}$ & $1.116_{\pm 0.032}$ \\
NetEst & $0.189_{\pm 0.080}$ & $0.490_{\pm 0.140}$ & $0.678_{\pm 0.200}$ & $0.138_{\pm 0.045}$ & $0.918_{\pm 0.103}$ & $1.059_{\pm 0.137}$ \\
DANN & $0.120_{\pm 0.032}$ & $1.048_{\pm 0.162}$ & $1.165_{\pm 0.171}$ & $0.128_{\pm 0.027}$ & $1.007_{\pm 0.038}$ & $1.136_{\pm 0.045}$ \\
IW-GCN & $0.131_{\pm 0.068}$ & $0.641_{\pm 0.134}$ & $0.774_{\pm 0.177}$ & $0.109_{\pm 0.024}$ & $0.946_{\pm 0.062}$ & $1.057_{\pm 0.069}$ \\
OM & \underline{$0.057$}$_{\pm 0.044}$ & \underline{$0.434$}$_{\pm 0.202}$ & \underline{$0.406$}$_{\pm 0.219}$ & $\bm{0.079}_{\pm 0.080}$ & \underline{$0.886$}$_{\pm 0.094}$ & $\bm{0.875}_{\pm 0.154}$ \\
TranCE (ours)                       & $\bm{0.051}_{\pm 0.047}$ & $\bm{0.381}_{\pm 0.205}$ & $\bm{0.354}_{\pm 0.204}$ & \underline{$0.080$}$_{\pm 0.080}$ & $\bm{0.879}_{\pm 0.098}$ & \underline{$0.876$}$_{\pm 0.155}$ \\
\bottomrule
\end{tabular}
\caption{Rotating-source transport error as absolute bias
$|\widehat{\mathbb E^{*}}[\tau^{R}]-\mathbb E^{*}[\tau^{R}]|$. Lower is
better, best per column in bold and second best underlined.}
\label{tab:source}
\end{table*}

\subsection{Baselines}

We compare against published network-effect and domain-transfer
estimators together with controlled ablations of our own pipeline.
\textbf{(a)} TARNet~\citep{shalit2017estimating}, trained on $\Pi$ and
applied to $\Pi^{*}$ with no network adjustment.
\textbf{(b)} OM~\citep{dahabreh2020extending}, the interventional
outcome model averaged over the $\Pi^{*}$ context without a
density-ratio correction, which is our own estimator with the transport
correction removed.
\textbf{(c)} IGL~\citep{sui2024invariant}, which learns graph
representations invariant across environments.
\textbf{(d)} Hoshino~\citep{hoshino2025transfer}, which
transports across domains by reweighting on the covariate marginal.
\textbf{(e)} IPW transport~\citep{horvitz1952generalization}, a
Horvitz--Thompson estimator that drops the outcome model and reweights
the $\Pi$ outcomes only.
\textbf{(f)} NetEst~\citep{jiang2022estimating}, a single-stream graph
outcome model applied to $\Pi^{*}$ with no correction.
\textbf{(g)} DANN~\citep{ganin2016domain}, whose graph encoder is
trained with a gradient-reversal penalty to align the embeddings of the
two domains.
\textbf{(h)} IW-GCN, a graph convolutional network~\citep{kipf2017semi}
whose loss on $\Pi$ is reweighted by the density ratio, the standard
importance-weighting response to covariate
shift~\citep{shimodaira2000improving}.

\subsection{Metrics}

For each effect
$R\in\{\mathrm{dir},\mathrm{spill},\mathrm{total}\}$ we report
absolute bias $|\widehat{\mathbb E^{*}}[\tau^{R}] - \mathbb E^{*}[\tau^{R}]|$ on the $\Pi^{*}$
average effect. For inference we report per-effect coverage of the $95\%$
bootstrap-with-refit interval together with its half-width, since
coverage alone can be bought by a wider interval. We repeat each experiment over
multiple random seeds and report the mean and standard deviation, with
the seed counts per experiment given in Appendix~C.

\subsection{Main Results}

TranCE attains the lowest bias on every Twitch effect and the lowest
spillover bias on both families. On Facebook-100 the uncorrected
substitution baseline OM is marginally ahead on the direct and total
effects, where the transport correction gives no measurable gain.
Table~\ref{tab:source} reports the transport error with every region
and school rotating through the role of interventional domain $\Pi$,
averaged over all within-family ordered pairs under strong covariate
shift. TARNet ignores the network and IPW carries no outcome model,
and both fall far behind on spillover, so both ingredients are
essential, while the graph-aware baselines improve on them without
adapting to $\Pi^{*}$.

\subsection{Ablations}
\label{sec:ablation}

To isolate the contribution of each component, we ablate one at a time
on the Twitch transport from DE to ES, FR, and PTBR, following the
cross-region setting of \citet{sui2024invariant}, with results in
Table~\ref{tab:ablation}. The transport correction carries the gain on
the two effects that depend on the neighbours, cutting the spillover
bias by $19\%$ and the total by $22\%$, and the H\'ajek
self-normalisation supplies about half of that. Neither helps the
direct effect, which the randomized own-treatment assignment already
identifies: the correction moves it by $0.0002$ on average, so the
component that repairs the spillover neither helps nor meaningfully
harms the effect that does not need repairing. Two neighbour treatment
levels lose resolution and cost $16\%$ on the spillover, while five
perform the same as three, so the map only has to be fine enough to
separate low from high exposure.

\begin{table}[t]
\centering

\scriptsize
\setlength{\tabcolsep}{2pt}
\begin{tabular}{lccc}
\toprule
 & dir & spill & total \\
\midrule
TranCE                         & $0.039_{\pm 0.028}$ & $\bm{0.253}_{\pm 0.070}$ & $\bm{0.221}_{\pm 0.087}$ \\
w/o transport correction       & $0.039_{\pm 0.032}$ & $0.313_{\pm 0.058}$ & $0.284_{\pm 0.075}$ \\
w/o H\'ajek self-normalisation & $0.039_{\pm 0.030}$ & $0.276_{\pm 0.069}$ & $0.245_{\pm 0.086}$ \\
2-level neighbour treatment map & $0.037_{\pm 0.028}$ & $0.294_{\pm 0.057}$ & $0.262_{\pm 0.053}$ \\
5-level neighbour treatment map & $\bm{0.035}_{\pm 0.023}$ & $0.256_{\pm 0.091}$ & $0.241_{\pm 0.088}$ \\
\bottomrule
\end{tabular}
\caption{Component ablations on the Twitch transport from DE to ES,
FR, and PTBR, absolute bias. Lower is better, best per column in bold.}
\label{tab:ablation}
\end{table}

\subsection{Double Robustness to a Misspecified Outcome Model}
\label{sec:dr-exp}
Table~\ref{tab:dr-full} shows the misspecification grid, in which the
two-stream model is replaced by one that drops the neighbour treatment
input, making its substitution spillover contrast structurally zero.
The correction improves the estimate in every arm where the
propensities are correct, but under the misspecified outcome model the
repair is partial, leaving a spillover bias of $0.964$ against the
$0.295$ it reaches when the outcome model is right. This is not a counterexample to Theorem~\ref{thm:dr},
whose guarantee is conditional on the overlap of
Assumption~\ref{ass:overlap}, and the diagnostics in Appendix~F show
that overlap is exactly what fails here rather than the stabilisation:
the shrinkage keeps essentially the whole correction in that arm, while
the target concentrates most of its nodes at the high neighbour
treatment level that the source under-populates, so the clipped weights
cannot fully stand in for the missing part of the target law. Since the randomized source has a known propensity,
outcome-model misspecification is the realistic failure mode in
deployment, so the substitution model is the component to get right. In
the adversarial rows the shrinkage falls back to the substitution under
a correct outcome model, and with both nuisances broken the estimator
fails as expected.

\begin{table}[t]
\centering

\scriptsize
\setlength{\tabcolsep}{3pt}
\begin{tabular}{llccc}
\toprule
Outcome model & Propensity & dir & spill & total \\
\midrule
Correct      & Correct & $0.036_{\pm 0.025}$ & $0.295_{\pm 0.071}$ & $0.272_{\pm 0.083}$ \\
Misspecified & Correct & $0.041_{\pm 0.029}$ & $0.964_{\pm 0.038}$ & $0.954_{\pm 0.054}$ \\
Correct      & Broken  & $0.037_{\pm 0.027}$ & $0.357_{\pm 0.058}$ & $0.331_{\pm 0.071}$ \\
Misspecified & Broken  & $0.040_{\pm 0.028}$ & $1.224_{\pm 0.050}$ & $1.202_{\pm 0.064}$ \\
\bottomrule
\end{tabular}
\caption{Double robustness on the Twitch transport from DE: absolute
bias per effect with the outcome model and the propensities each
correct or broken. Lower is better.}
\label{tab:dr-full}
\end{table}

\subsection{Where Transport Is Hard}
\label{sec:multisource}
\begin{figure}[t]
  \centering
  \includegraphics[width=0.78\columnwidth]{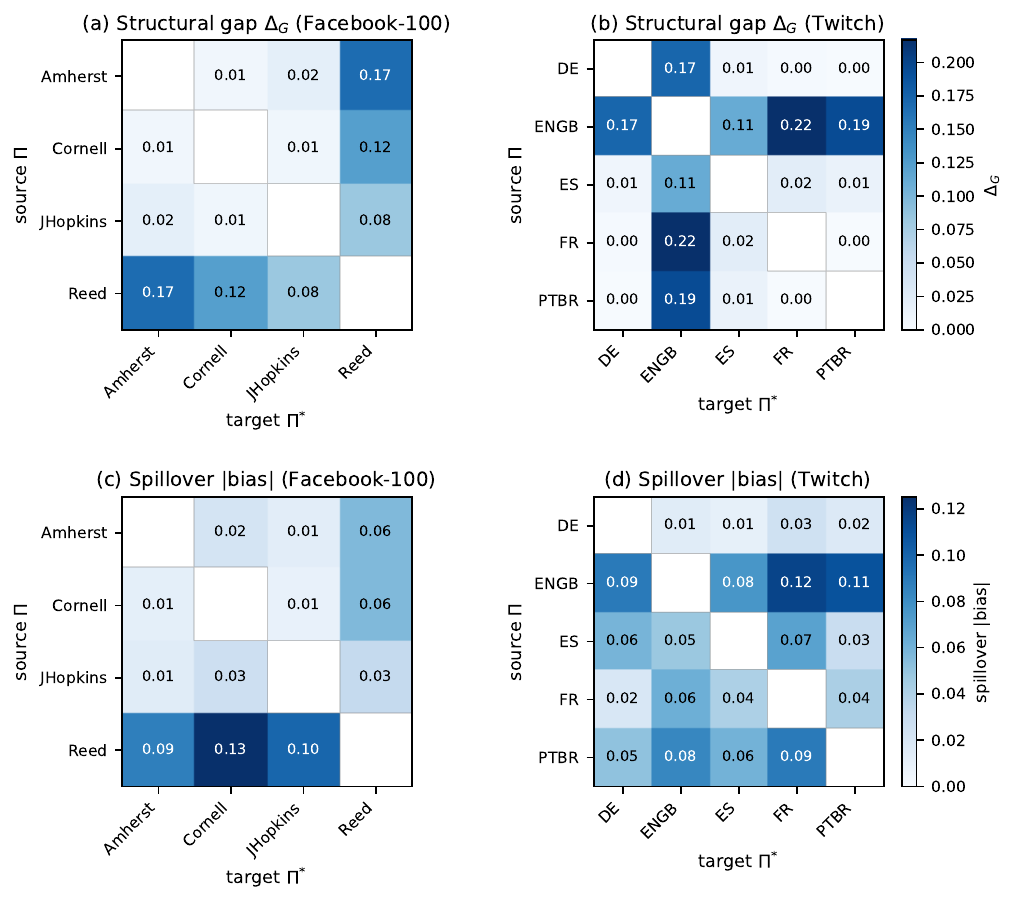}
  \caption{Structural gap $\Delta_G$ (top) and transported spillover
  bias (bottom) across all within-family source-target pairs. The two
  panels in each row share one colour scale.}
  \label{fig:multisource}
\end{figure}

Figure~\ref{fig:multisource} shows the structural gap $\Delta_G$ and
the transported spillover bias across all within-family source-target
pairs, with the covariate shift switched off so the structural gap is
the only thing moving. Transport accuracy degrades with the structural
distance between the two networks: the two quantities correlate at
Pearson $r=0.66$ on Twitch and $r=0.64$ on Facebook-100, and the pairs
involving a structural outlier, ENGB and Reed, carry both the largest
gap and a mean spillover bias of $0.075$ against $0.034$ for ordinary
pairs. Since
$\Delta_G$ needs only the two graphs and no outcomes, the error is
predictable before deployment, and a large gap is a signal to widen the
reported interval or to seek a closer source.

\subsection{Sensitivity to Hidden Confounding}
\label{sec:hidden}

To probe the sensitivity to hidden confounding, we add a hidden
structural confounder of strength $\gamma$ to the outcome through the
standardized neighbour-mean degree, a quantity withheld from
$\bm w$, with the covariate shift zeroed so $\gamma$ is the only
moving part. The spillover bias grows steadily with $\gamma$ while the
direct effect stays flat, and the uncorrected substitution estimator
tracks TranCE throughout, since a confounder outside the adjustment set
violates the condition of
Theorem~\ref{thm:transport} and is invisible to the outcome model and
the density ratio alike.

\subsection{Real-World Validation on a Field Experiment}
\label{sec:casestudy}

Table~\ref{tab:casestudy} shows the leave-region-out validation on the
weather-insurance field experiment of \citet{cai2015social} in rural
China, which randomized an information session at the household level
across three separated regions and recorded insurance take-up on a real
friendship network. TranCE agrees with the held-out region's own
randomized estimate in five of six comparisons. Each region being itself
randomized, a held-out region is a within-experiment gold standard: we
rotate it over all three, transport both effects from the other two,
and compare against the region's own randomized estimate, with
village-cluster bootstrap intervals that refit every nuisance. The
direct effect matches the gold standard closely and covers in all three
regions,
while the spillover covers in two of three, the exception being the
region whose own within-experiment spillover is near zero. The
spillover is identified only up to the treated-friend share being as
good as random given degree, so it transports less precisely and with
wider intervals. This is direct evidence that the estimator carries
real experimental effects across real networks.

\begin{table}[t]
\centering

\scriptsize
\setlength{\tabcolsep}{1.35pt}
\begin{tabular}{lcccccc}
\toprule
 & \multicolumn{3}{c}{Direct} & \multicolumn{3}{c}{Spillover} \\
\cmidrule(lr){2-4}\cmidrule(lr){5-7}
Held-out & Est. & Gold & $95\%$ CI & Est. & Gold & $95\%$ CI \\
\midrule
Region 1 & $0.085$ & $0.076$ & $[0.049,0.114]$\,\checkmark
         & $0.036$ & $0.048$ & $[\text{-}0.035,0.087]$\,\checkmark \\
Region 2 & $0.068$ & $0.078$ & $[0.027,0.092]$\,\checkmark
         & $0.056$ & $-0.007$ & $[0.004,0.090]$\,$\times$ \\
Region 3 & $0.056$ & $0.066$ & $[0.026,0.082]$\,\checkmark
         & $0.015$ & $0.030$ & $[\text{-}0.014,0.058]$\,\checkmark \\
\bottomrule
\end{tabular}
\caption{Leave-region-out validation. Gold is the held-out region's own
randomized estimate, and \checkmark{} marks the $95\%$ bootstrap
interval containing it.}
\label{tab:casestudy}
\end{table}

\section{Conclusion}
\label{sec:conclusion}

We gave an algorithm for transportable causal effect estimation on
network data, employing a selection diagram in the network setting to
separate covariate shift from structural shift, deriving transport
formulas for the direct, spillover, and total effects, and implementing
the formulas with a doubly-robust algorithm with cross-fitted
inference.
We validated the estimator on the Twitch and Facebook-100 families with
every domain rotating as the source, and on a real weather-insurance
field experiment whose transported effects match held-out randomized
gold standards. One
limitation is that the neighbour treatment map summarises neighbourhood treatment
through a small set of fixed levels. Extending it to richer neighbour-treatment summaries is natural future
work.

\bibliography{references}

\appendix

\end{document}